\documentclass[amsmath,amssymb,aps,reprint,prx,showkeys]{revtex4-2}
\usepackage{graphicx}
\usepackage{mathtools}
\usepackage{subfigure,array}
\usepackage{hyperref}
\usepackage[english]{babel}

\hypersetup{
 colorlinks  = true,
 urlcolor   = blue, 
 linkcolor  = blue, 
 citecolor  = blue 
}
\usepackage{float}
\newcolumntype{C}[1]{>{\centering\let\newline\\\arraybackslash\hspace{0pt}}m{#1}}
\begin{abstract}
Reservoir computers (RCs) provide a computationally efficient alternative to deep learning while also offering a framework for incorporating brain-inspired computational principles. By using an internal neural network with random, fixed connections—the ‘reservoir’—and training only the output weights, RCs simplify the training process but remain sensitive to the choice of hyperparameters that govern activation functions and network architecture. Moreover, typical RC implementations overlook a critical aspect of neuronal dynamics: the balance between excitatory and inhibitory (E-I) signals, which is essential for robust brain function. We show that RCs characteristically perform best in balanced or slightly over-inhibited regimes, outperforming excitation-dominated ones. To reduce the need for precise hyperparameter tuning, we introduce a self-adapting mechanism that locally adjusts E/I balance to achieve target neuronal firing rates, improving performance by up to 130\% in tasks like memory capacity and time series prediction compared with globally tuned RCs. Incorporating brain-inspired heterogeneity in target neuronal firing rates further reduces the need for fine-tuning hyperparameters and enables RCs to excel across linear and non-linear tasks. These results support a shift from static optimization to dynamic adaptation in reservoir design, demonstrating how brain-inspired mechanisms can improve RC performance and robustness while deepening our understanding of neural computation. 

\end{abstract}

\begin{document}
\makeatletter
\title{Boosting Reservoir Computing with Brain-inspired Adaptive Dynamics}
\author{Keshav Srinivasan}
\affiliation{Biophysics Program, University of Maryland, College Park, MD 20740, USA}
\affiliation{Section on Critical Brain Dynamics, National Institute of Mental Health, Bethesda, MD 20892, USA}
\author{Dietmar Plenz}
\affiliation{Section on Critical Brain Dynamics, National Institute of Mental Health, Bethesda, MD 20892, USA}
\author{Michelle Girvan}
\affiliation{Biophysics Program, University of Maryland, College Park, MD 20740, USA}
\affiliation{Department of Physics, University of Maryland, College Park, MD 20740, USA}
\affiliation{Santa Fe Institute, Santa Fe, NM 87501, USA}

\maketitle

\section{Introduction}
Reservoir computers \cite{jaeger_echo_2001,maass_real-time_2002} (RCs) offer a computationally efficient framework for implementing recurrent neural networks (RNNs), broadly reflecting key principles of the brain’s neocortex. RNNs process data through recurrent connections that generate complex activity patterns, making them well-suited for dynamic computations. RCs harness this feature using a three-layer architecture, where a fixed-weight, randomly connected RNN, the reservoir, transforms low dimensional input into high-dimensional internal representations. A trainable output layer then maps these representations to task-specific outputs. A key strength of RCs lie in their training efficiency; by keeping the reservoir fixed and training only the output layer, RCs achieves faster training and lower computational demand \cite{soriano_minimal_2015,sakemi_learning_2024,cisneros_benchmarking_2022} compared with other machine learning approaches, such as convolutional neural networks (CNNs), long-short term memory networks (LSTMs), and transformers \cite{wei_virvision_2021}. These alternatives typically rely on feed-forward architectures requiring extensive multiple-layer training via back-propagation \cite{lillicrap_backpropagation_2020}and substantial data and computational resources.

The random yet fixed initialization of the reservoir poses a challenge for optimizing RC performance. This initialization determines the internal dynamics of the reservoir, which directly influence performance, but there is little guidance on how to systematically construct effective reservoirs. Theoretical studies \cite{carroll_network_2019, carroll_reservoir_2020} suggest that optimal performance may occur near the edge of chaos, indicating that maintaining a delicate dynamical balance within the reservoir is beneficial for effective design. This need for a delicate dynamical balance in RCs parallel the role of excitatory (E) and inhibitory (I) balance in RNNs of the brain’s neocortex, which is crucial for optimal brain function \cite{dehghani_dynamic_2016}. Disruptions in this balance are linked to altered states of consciousness, such as anesthesia \cite{maclver_volatile_1996}, coma, and depression \cite{hamani_subcallosal_2011}, which correspond to over-inhibited states, while conditions like epilepsy arise from excessive excitation \cite{yizhar_neocortical_2011}.In neuronal network models, deviations from optimal E-I balance impair key aspects of information processing, including dynamic range and information capacity \cite{shew_neuronal_2009}. 

Traditional RCs capture some of this dynamical balance by including random positive as well as negative connections between neurons in the reservoir. However, the effects of changing the relative balance of excitation and inhibition remains underexplored. Further, in the neocortex, neurons and their connections have distinct excitatory and inhibitory roles, following Dale’s Law \cite{kandel_cellular_1968, strata_dales_1999}, where each neuron maintains only excitatory or inhibitory outgoing links. To bridge this gap, we introduce a brain-inspired RC architecture with distinct excitatory and inhibitory roles for neurons and their connections. Our design maintains a 4:1 ratio in the number of excitatory to inhibitory neurons and begins with a base structure that achieves an overall global E/I balance through a 1:4 ratio in the strengths of excitatory and inhibitory links, consistent with their approximate relative proportions in the brain \cite{somogyi_salient_1998}. By explicitly incorporating tunable E/I balance through adjustments to inhibitory link strengths, we enable RCs to achieve better performance across tasks while adhering to fundamental principles of neural organization.

Specifically, we explore E-I balance by incorporating a local plasticity rule that adapts inhibitory weights to achieve target firing rates, inspired by activity homeostasis in neurobiology \cite{prinz_similar_2004, turrigiano_homeostatic_2004, turrigiano_self-tuning_2008, pozo_unraveling_2010, davis_homeostatic_2006}. Unlike traditional RCs, which have fixed global balance from random network construction, this autonomous adaptation allows adjustable local and global E-I balance, improving performance across diverse tasks. Typical RC hyperparameter tuning is computationally expensive \cite{joy_rctorch_2022}, whereas our biologically inspired approach enables autonomous self-adaptations that efficiently tune internal dynamics for broad applicability. While previous unsupervised learning approaches have investigated plasticity mechanisms in recurrent networks \cite{del_papa_criticality_2017,lazar_sorn_2009,triesch_synergies_2007,zenke_plasticity_2011}, and RCs \cite{babinec_improving_2007,triesch_synergies_2007,morales_unveiling_2021,yusoff_modeling_2016} specifically, none have directly focused on tuning E-I balance. By introducing heterogeneous target rates, our model reduces the need for fine-tuning RC performance across linear tasks, such as memory recall, and nonlinear time-series prediction tasks. Additionally, we demonstrate that, relying on iterative tuning, networks can be designed to achieve desired levels of local balance through a one-step adjustment of randomly initialized inhibitory links, further improving efficiency and scalability crucial for large RCs.

Our work builds on prior research\cite{soriano_minimal_2015, yamazaki_spiking_2007, schmitt_efficient_2023, sumi_biological_2023, cucchi_reservoir_2021}exploring the shared properties between biological neural networks and reservoir computers, particularly regarding the role of inhibition in these networks \cite{enel_reservoir_2016, ju_spatiotemporal_2015}. By demonstrating how local tuning of E-I balance enhances RC performance, our results bridge neuroscience and artificial intelligence, leveraging biological insights to improve machine learning.

\section{Methods}
\subsection{Excitatory-Inhibitory (E-I) Reservoir Computing}
A reservoir computer is a type of recurrent neural network (RNN) structured into three layers: the input layer, the reservoir, and the output layer (See Fig. 1A). Neuron-like units in the reservoir are connected by fixed, random links (including feedback loops). Training occurs exclusively in the output layer, leaving the connection strengths in both the input layer and the reservoir unchanged in the learning process.

Traditional RCs use random link weights but lack distinct excitatory and inhibitory nodes or neurons, limiting systematic analysis of E-I balance in analogy to brain dynamics. Our framework introduces a tunable E-I reservoir that utilizes sigmoidal activation functions (see below) to ensure non-negative neuronal states, removing ambiguity in the roles of excitation and inhibition. This guarantees that positive link weights are always excitatory, while negative link weights are always inhibitory. A similar effect could be achieved by shifting the commonly used ‘tanh’ activation (e.g., by adding 1) to restrict reservoir neurons to non-negative values. Each neuron's state at any given time is determined by two primary terms: 

\textit{Membrane potential} (\( V\)): This internal variable describes the state of a neuron based on all its influencing factors.

\textit{Firing variable} (\( r\)): This external variable represents the observable firing rate state of a neuron, dependent on the membrane potential and the neuron's activation threshold. Values close to 0 represent no firing, and values close to 1 represent saturated firing.
The following equations govern the dynamics of these two variables:
\begin{equation*}
V_{i}\left(t+\Delta t\right) = \lambda_{i}V_{i}\left(t\right)+\sum_{j = 1}^{N}\left(A_{ij}^{E}-A_{ij}^{I}\right)r_{j}\left(t\right)+W_{i}^{in}u\left(t\right)
\end{equation*}
\begin{equation}
r_{i}\left(t\right) =\text{Sig}\left(V_{i}\left(t\right)-\theta_{i}\left(t\right)\right)
\end{equation}
Here, \( \lambda_{i}\) is the leakage constant of neuron \( i\), representing how much of its previous state is retained at each time step, and the reservoir time step \( \Delta t\) is set to match the sampling time of the input \( u\left(t\right).\) In our study, \( \lambda_{i}\) is set to 0, meaning individual neurons have no memory of their previous state, and the system’s memory, which emerges from recurrent interactions, is encoded in the full reservoir state. The weighted connectivity matrix, \( A = A^{E}-A^{I}\), defines how neurons in the network are connected. The network structure adheres to Dale’s law, where excitatory neurons have only excitatory outgoing links, and inhibitory neurons have only inhibitory outgoing links. A fraction of the neurons are designated as excitatory, with their outgoing synapses making up the excitatory submatrix \( A^{E}\), and the remaining neurons are labeled inhibitory, with their outgoing synapses making up the inhibitory submatrix \( A^{I}\) (both non-negative; see Appendix for details on the construction of the adjacency matrix). 

The global relative E-I balance is controlled by a global balance parameter, \( \beta\), which represents the relative total strength of the excitatory and inhibitory interactions. This parameter is defined as:

\begin{equation}
\beta =\left\langle \beta_{i} = \sum_{j}^{}A_{ij}^{E}-A_{ij}^{I}\right\rangle
\end{equation}

where \( \beta_{i}\) is the local balance of neuron \( i\). In our studies, if \( \beta = 0\), the network is balanced, but \( \beta\) can range from very negative (inhibition dominated) to very positive (excitation dominated). We vary this E-I balance by varying the strength of inhibitory links only, keeping other parameters fixed. The firing rate of each neuron, \( r_{i}\), is determined by the threshold, \( \theta_{i}\), and the activation function of each neuron (Fig.\ \hyperref[Fig1]{1A} \textit{inset}). In our setup, the activation function is a sigmoid function, given by \(\text{Sig}\left(x\right) = 1/\left(1+e^{-cx}\right),\) This model approximates a spiking network as \( c \rightarrow \infty\) and linear activation as \( c \rightarrow 0\).

The input is fed into the reservoir via the input layer, which links the input signal to a fraction, \( f_{in}\), of the reservoir neurons. The non-zero values in the input matrix, \(\mathbf{W}_{\mathbf{in}}\)\, are drawn from a uniformly distributed random variable within the interval [-\( \sigma_{in}/2\), \( \sigma_{in}/2\)], where \( \sigma_{in}\) is the input link spread. The output layer, \(\mathbf{W}_{\mathbf{out}}\), functions as a readout, with weights connecting to each neuron in the reservoir. These weights are trained using ridge regression with a regularization parameter, \( \eta\). Unless stated otherwise, the default values for the parameters used in this study are provided in Table \ref{Table1}.

\begin{table}
    \centering
\begin{tabular}{|c|c|c|}
 \hline
\textbf{Description}             & \textbf{Symbol}            & \textbf{Value}\\
\hline
Number of neurons           & $N$           & 500\\
Excitatory fraction         & $f_E$         & 0.8\\
Inhibitory fraction         & $f_I$         & 0.2 \\
Average network degree      & $k_E=k_I$             & 50\\
Mean exc. synaptic strength & $\mu_E$ & $\left(\frac{1}{k\cdot f_E}\right)=\frac{1}{40}$\\
Exc. synaptic spread          & $\sigma_E$          & $0.2\mu_E$\\
Mean inh. synaptic strength         & $\mu_I$ & $-\left(\frac{f_E}{f_I}\right)\mu_E$*\\
Balance parameter       & $\beta$           & 0*\\
Inh. synaptic spread    & $\sigma_I=\sigma_E $         & $0.2\mu_E$\\
Scaling factor        & $\alpha$               & 1\\
Threshold          & $\theta$          & 0*\\
Sigmoid steepness          & $c$          & 10\\
Input fraction & $f_{in}$           & 0.3\\
Input link spread & $\sigma_{in}$           & Fig.\ \ref{Fig5}*\\
Leakage constant & $\lambda_i$           & 0\\
Regularization parameter & $\eta$           & $10^{-7}$\\
\hline
\end{tabular}
\caption{(* indicates tunable parameters). The total neuron count, $N$, is split into excitatory and inhibitory populations based on the excitatory fraction, $f_E$ ($f_I=1-f_E$). The connectivity matrix has an average network degree, $k_E=k_I=50$. Excitatory synapse strengths follow a Gaussian distribution with mean $\mu_E$ and standard deviation $\sigma_E$. Inhibitory synapse strengths have a mean, $\mu_I$, which can be varied to achieve the desired balance parameter, $\beta$ (see Eq. 10) and a variance $\sigma_I=\sigma_E$. In balanced ($\beta=0$) networks, $\mu_I$ is set to compensate for different fractions of excitatory and inhibitory neurons. A scaling factor, $\alpha$, scales all elements of the connectivity matrix.}
\label{Table1}
\end{table}
\subsection{Neuron differential entropy}
In order to quantify reservoir dynamics, we calculate the neuron-level entropy, previously linked to RC performance \cite{hurley_tuning_2023, ozturk_analysis_2007}. Instead of system-wide entropy, we compute individual neuron entropy using a non-parametric differential entropy estimator \cite{berrett_efficient_2019}, avoiding binning issues (see Appendix C for details). Although neuronal entropy is associated with performance, high entropy does not guarantee good performance, as high-entropy states may still be entirely uncorrelated with the input.

\subsection{Tasks for testing}
We assess the performance of our reservoir computer via a set of well-known tasks, designed to test different computational challenges. We consider the following 4 tasks:

\textit{Memory Capacity task:} A widely used memory benchmark task, used by Jaeger (2002), involves feeding the reservoir a random signal, \( u\left(t\right)\) with values uniformly sampled from the interval [0,1]. The goal of this task is to train the reservoir to recall a value it encountered \( d\) steps earlier. For each delay, \( d\), the accuracy of the recall is measured using a coefficient of determination, \( R^{2}\), given by: 
\begin{equation}
R^{2} = \frac{cov\left(u\left(t-d\right),y_{d}\left(t\right)\right)^{2}}{var\left(u\left(t\right)\right)\cdot var\left(y_{d}\left(t\right)\right)}
\end{equation}
Here \( u\left(t\right)\) represents the input and \( y_{d}\) denotes the reservoir output for a given delay, \( d\). A memory capacity curve (Fig.\ \hyperref[Fig1]{1B}, i) shows the plot of \( R^{2}\) as a function of delay. The overall memory capacity is determined by calculating the total area under this curve, approximated by summing the values of \( R^{2}\). As a practical constraint, this summation is capped at \( d = 70\).

\textit{ NARMA-10 Task:} Another commonly used benchmark for evaluating a system's computational performance is the Nonlinear Autoregressive Moving Average (NARMA) test \cite{jaeger_echo_2001,wang_deep_2021}, which is a discrete temporal task with a 10th-order lag. The formulation for a NARMA-10 time series is: 
\begin{align}
y\left(t\right) &= py\left(t-\Delta t\right)+qy\left(t-\Delta t\right)\sum_{i = 1}^{10}y\left(t-i\Delta t\right) \nonumber\\ 
&+g\cdot u\left(t-10\Delta t\right)u\left(t-\Delta t\right)+D  
\end{align}
where the parameters are defined as \( p = 0.3\), \( q = 0.05\), \( g = 1.5\), \( D = 0.1\). The stochastic input \( u\left(t\right)\) is sampled from a uniform distribution over the range [0, 0.5] with timestep \( \Delta t = 1\). This test is designed to challenge the system's ability to handle nonlinearities as well as its capacity to retain memory across substantial time lags. Performance was evaluated by calculating the root mean squared error (RMSE) between the value given by Equation 5 and the trained reservoir output. 

\textit{Mackey-Glass Time-series Prediction:} Chaotic time-series prediction is an area where RCs have been shown to be particularly advantageous \cite{pathak_model-free_2018, jaeger_harnessing_2004}. By drawing from a repository of possible dynamics, once trained, RCs faithfully model the underlying behavior of chaotic systems, predicting system behavior for extended periods before deviations occur due to the accumulation of small errors, a characteristic of chaotic dynamics. In this context, the Mackey-Glass system \cite{mackey_oscillation_1977} is a widely used chaotic model frequently employed to evaluate RC performance. It is described by the following one-dimensional delay differential equation: 
\begin{equation}
\frac{dx}{dt} = \xi\frac{x\left(t-\tau\right)}{1+x^{n}\left(t-\tau\right)}-\gamma x  
\end{equation}
The specific parameter values used to generate chaotic behavior are \( \xi = 0.2\), \( \gamma = 0.1\), \( \tau = 17\), and \( n = 10\). The time series was computed using the fourth-order Runge-Kutta method with integration timestep equal to the sampling timestep \( \Delta t = 0.1s\). Performance is quantified as the valid prediction time (VPT, defined as the first time point at which the normalized root mean squared error (NRMSE) between the predicted state and true state (using Eqn. 6) exceeds a threshold (with a default value of 0.4).

\textit{  Partially Observed Lorenz Time-Series Prediction:} A common test for chaotic time series prediction in RCs is the Lorenz system, a set of three differential equations modeling atmospheric convection \cite{lorenz_deterministic_1963}, known for its chaos under certain parameters. The equations defining the Lorenz system are \begin{align}
\dot{x} &= a\left(y-x\right) \nonumber\\ 
\dot{y} &= -xz+bx-y\\ 
\dot{z} &= xy-cz \nonumber
\end{align}
Where \( x\), \( y\), and \( z\) represent the internal state variables, each evolving according to the given set of differential equations. The parameter values \( a = 10\), \( b = 28\), and \( c = 8/3\) are known to generate chaotic dynamics. The time series were computed using the fourth-order Runge-Kutta method with an integration timestep of \( \delta t = 0.01s\) and a sampling timestep of \( \Delta t = 0.02s\). For our study, we focus on one of the system variables, \( x\), and keep the other 2 hidden from the RC. This approach models scenarios where only partial measurement of system state is available, relevant for real-world applications. To facilitate comparability, Mackey-Glass and Lorenz system values are normalized between 0 and 1.

\subsection{Inhibitory adaptation mechanism}
We introduce an inhibitory adaptation mechanism that drives neurons toward a desired level of local balance (Fig.\ \hyperref[Fig2]{2A}). Neurons exceeding their target firing rate strengthen inhibitory synapses, while those below reduce inhibition. This computationally low-cost self-tuning process adjusts only internal reservoir weights, shaping intrinsic dynamics prior to more costly task-specific training of the output layer. The adaptation of inhibitory connections is governed by the following equation:
\begin{equation}
 A_{ij}^{I}\left(t+1\right) = A_{ij}^{I}\left(t\right)+\delta\left(r_{i}\left(t\right)-\rho_{i}\right)
\end{equation}
In this context, \( A_{ij}^{I}\geq 0\) denotes strength of the inhibitory link from \( j\) to neuron \( i\), \( \delta\) is the learning rate which in our case is set to \( 10^{-3}\), \( r_{i}\) represents the state variable of neuron \( i\), and \( \rho_{i}\) is the target firing rate of neuron \( i\). 
We examine two scenarios: (1) a homogeneous target firing rate for all neurons and (2) a heterogeneous distribution of target rates. A target rate of \( \rho_{T}\) $=$ 0.5 represents locally balanced excitation and inhibition, while values above or below indicate over-excitation or inhibition. This parameter is central to our adaptation process. We begin by exploring the homogeneous case in which all neurons have the same target firing rate and then consider heterogeneity in the neuronal target firing rates, which is designed to reflect neuronal heterogeneity in the brain and also has the added benefit of eliminating the necessity for fine-tuning the target firing rate using the beta distribution (Fig.\ \ref{Fig3}; see Appendix B).
\subsection{Designing reservoirs with controlled levels local E-I balance}
While our inhibitory adaptation rule relies only on local information to adjust synapses—an approach inspired by biological systems—full knowledge of the system and its synaptic weights allows direct design of a reservoir with controlled local E-I balance by adjusting inhibitory synapses to achieve the target firing-rate distribution. Our designed network starts as before with a randomly configured network, but instead of small, iterative adjustments, we compute a one-step rescaling to achieve the target firing rate, \( \rho_{i}\). Each neuron receives a multiplicative factor (\( \Omega_{i}\)) applied to its inhibitory synapses, significantly accelerating the adaptation process to a desired local balance. This factor is calculated as:

\begin{equation}
\Omega_{i} =\frac{\text{Sig}^{-1}\left(\rho_{i}\right)+\theta_{i}-W_{i}^{in}\left\langle u\right\rangle -\sum A_{ij}^{E}\rho_{j}}{\sum A_{ij}^{I}\rho_{j}} 
\end{equation}
where Sig\textsuperscript{-1} function is an inverse sigmoid function, \(\left\langle u\right\rangle\) is the mean input and \( \rho_{j}\) is the target firing rate of neuron \( j\) (see Supplementary Material for details).
\begin{figure*}
\includegraphics[width=16.3cm]{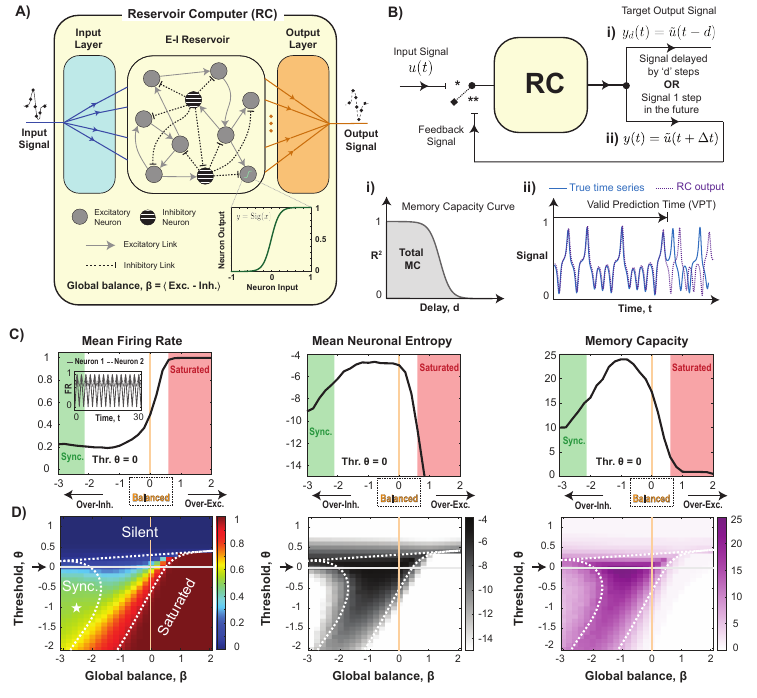}
\caption{\textbf{Balanced to slightly inhibited reservoirs exhibit robust, high performance devoid of synchronized or saturation dynamics.} \textbf{(A)} Three-layered schematic of the E-I RC architecture. Excitatory (\textit{solid gray}) and inhibitory (\textit{striped}) reservoir neurons with sigmoid input-output function form excitatory (\textit{solid}) and inhibitory (\textit{dashed}) connections, respectively. \textbf{(B)} RC evaluation across tasks : i) Memory task - RC runs in open-loop ($\ast$) mode, receiving input $u(t)$ and predicting it $d$ steps later. Accuracy is measured by $R^2$, with memory capacity as the area under the $R^2$ vs. delay curve. ii) Other tasks - RC runs in open-loop (NARMA-10) or closed-loop (Mackey-Glass, Lorenz) mode ($\ast$$\ast$), predicting one step ahead. Accuracy is measured by RMSE or Valid Prediction Time (VPT). \textbf{(C)} RC dynamics and performance in the memory test for $\theta=0$. \textit{Left:} Mean firing rate, \textit{Middle:} Neuronal entropy and \textit{Right:} Memory capacity. Synchronized (inset, \textit{green shaded}) and saturated (\textit{red shaded}) regimes are highlighted. \textbf{(D)} RC dynamics and performance in the memory test with variable thresholds reveals dynamical landscape with three extreme regimes: silent ($\sim$0), saturated ($\sim$1), and globally synchronized ($\bar{C_{ij}}>0.9$, neither silent nor saturated). Neuronal entropy maps these regions, showing reduced entropy in all three. Memory capacity is highest in balanced/slightly inhibited states and declines in the over-excited regime.}
\label{Fig1}
\end{figure*}
\section{Results}
\subsection{Robust performance in balanced and slightly inhibited regimes, bounded by silent, saturated and synchronized regimes; brittleness with over-excitation}
We designed a tunable excitatory-inhibitory (E-I) reservoir with 400 excitatory and 100 inhibitory randomly connected neurons (10$\%$ connectivity; average network degree $=$ 50). In this randomly connected, sparse network, neuronal firing rates \( r_{i}\in [0,1]\) are determined by a sigmoidal activation function that transforms integrated inputs—representing the ‘membrane potential’ above a bias, or threshold, \( \theta\)—ensuring non-negative reservoir states, \(\mathbf{r}\left(t\right)\). The network structure follows Dale’s law with distinct excitatory and inhibitory roles for neurons (Fig.\ \hyperref[Fig1]{1A}). The fixed input layer, \(\mathbf{W}_{\mathbf{in}}\), projects the input signal, \( u\left(t\right)\), onto 30$\%$ of randomly selected excitatory and inhibitory reservoir neurons, and the trained output layer, \(\mathbf{W}_{\mathbf{out}}\), maps neuronal firing rates of all neurons to the output signal, \( y\left(t\right)\). Reservoir dynamics were quantified using the mean firing rate across all neurons and time, \(\left\langle\overline{r}\right\rangle_{t}\) and the neuronal state entropy, \( H\left(\mathbf{r}\right)\), a known correlate of RC performance \cite{hurley_tuning_2023, ozturk_analysis_2007}.

RC performance was evaluated for the four aforementioned tasks (Fig.\ \hyperref[Fig1]{1B}): a memory capacity task \cite{jaeger_echo_2001, rodriguez_optimal_2019}, which measures how long the reservoir can retain information about past inputs, the Nonlinear Autoregressive Moving Average (NARMA-10) task, which requires both memory and nonlinearity due to its 10th order lag, and the Mackey-Glass and Lorenz system, two chaotic time series prediction tasks, a known strength of RCs \cite{pathak_model-free_2018, jaeger_harnessing_2004}. 

We first discuss RC performance at threshold \( \theta = 0\) as the global balance parameter,\textit{ }\( \beta\), is changed by varying the mean inhibitory synapse strength\( , \mu_{I}\) (Eq. 10) but keeping the inhibitory neuron fraction fixed at \( f_{I} = 0.2\) and mean excitatory synapse strength, \( \mu_{E} = 0.025\). This shift in the global balance parameter, \( \beta\), defines distinct dynamical regimes: over-excited reservoirs (\( \beta >0.5\)) saturate rapidly (\(\left\langle\overline{r}\right\rangle_{t}>0.95\); Fig.\ \hyperref[Fig1]{1C}, \textit{red}), while over-inhibited reservoirs (\( \beta <-2\)) exhibit low entropy, globally synchronized oscillations (with high mean pairwise correlations: \(\overline{C_{ij}}>0.9\); neither silent or saturated dynamics: \( 0.05<\left\langle\overline{r}\right\rangle_{t}<0.95\); Fig.\ \hyperref[Fig1]{1C} inset, \textit{green}). High RC performance and entropy occur only between these extremes, with maximal outcome robustly placed in the slightly inhibited to balanced regime (\( -2<\beta \leq 0\)). Similar trends are observed when global balance is varied through alternative means—for example, by increasing the inhibitory neuron fraction to $f_I=0.5$  or by keeping synaptic strengths fixed and adjusting the proportion of inhibitory neuron (Supp. Fig. 1). This robust performance pattern holds across all four tasks, not just memory capacity (Supp. Fig. 2). We also note that restricting outgoing inhibitory connections to one category of neurons (Dale’s law) does not improve or impair RC performance; we see that shuffling the connections yields similar results (Supp. Fig. 3) 
\vspace{-0.01cm}

Across both positive and negative thresholds, the reservoir dynamics vary significantly with global balance (Fig.\ \hyperref[Fig1]{1D}). For networks with $\beta>1$, the mean inhibitory synaptic strength, $\mu_I$, becomes negative (Eq. 9), meaning that the neurons that were initially labeled inhibitory will have excitatory effects. In these over-excited networks, the reservoir exhibits an abrupt transition from silence to saturation, making it highly sensitive to threshold changes. In contrast, networks with stronger inhibition exhibit extended regions of intermediate firing rates, supporting high entropy and performance. This broad high-performance band for \( \beta <0\) spans the transition between synchronized and saturating regimes. The center of this band shifts toward increased inhibition to counterbalance lower thresholds. However, even a modest increase in the threshold above zero rapidly eliminates this high-performance regime, pushing the reservoir into a silent state (\(\left\langle\overline{r}\right\rangle_{t}<0.05\)) regardless of inhibition levels.

\onecolumngrid\
\begin{figure}[H]
\centering
\includegraphics[width=16.5cm]{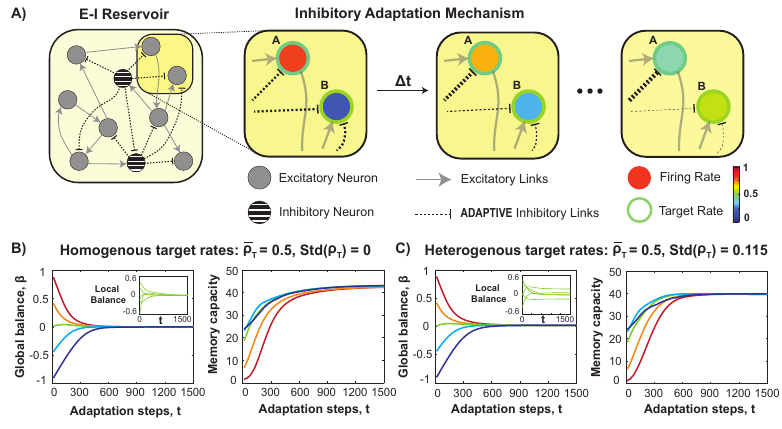}
\caption{\textbf{The inhibitory adaptation mechanism enhances both global and local balance, improving memory performance across homogeneous and heterogeneous target firing rates.} \textbf{(A)} Schematic of adaptation mechanism: inhibitory connections strengthen for neurons with firing rates (\textit{fill color}) above their target rate (\textit{border color}) and weaken for those below, restoring local balance.\textbf{ (B) }Evolution of global balance and memory capacity across adaptation steps for five initial conditions (ranging from over-inhibited: \textit{dark blue}, to over-excited: \textit{dark red}) with homogeneous target firing rates. All networks converge to a balanced state with improved memory. Even the initially globally balanced reservoir (\textit{green curve}) has improved memory capacity without a visible shift in global balance, as correcting large local imbalances (\textit{inset}) enhances performance. \textbf{(C) }Similar analysis with heterogeneous target firing rates (drawn from a beta distribution). The network converges to global balance and improved performance while preserving slight variations in local balance (\textit{inset}).}
\label{Fig2}
\end{figure}
\newpage
\begin{figure}[H]
\centering
\includegraphics[width=17cm]{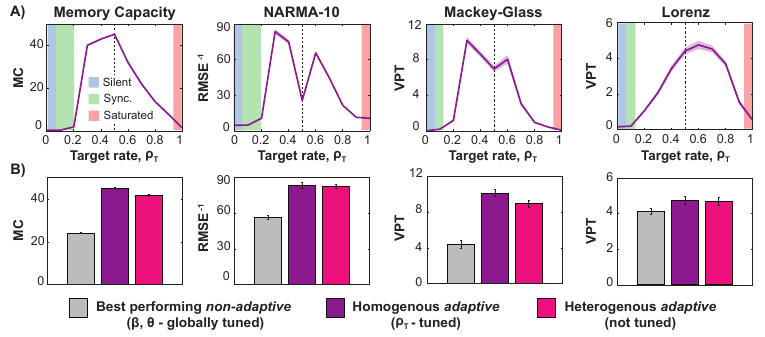}
\caption{\textbf{Adaptive reservoirs achieve optimal performance at different target firing rates for different tasks, significantly outperforming non-adaptive, globally tuned reservoirs. (A)} Optimal performance (\textit{purple line}: mean across 100 runs, \textit{shaded area}: standard error of the mean) varies with homogenous target firing rate ($\rho_T$). Linear tasks, such as memory capacity, peak near the linear midpoint of the sigmoid (\textit{dashed line}), while others such as NARMA-10, Mackey-Glass and Lorenz benefit from slight deviations above or below 0.5, leveraging nonlinear effects. \textbf{(B)} Performance of non-adaptive vs. adaptive reservoirs. \textit{Gray bars}: best non-adaptive, globally tuned RC. \textit{Colored bars}: adaptive reservoirs with tuned homogeneous (\textit{purple}) or fixed heterogeneous (\textit{pink}) target rates. Adaptive reservoirs significantly improve performance, with heterogeneous rates offering robust, task-general performance without requiring firing rate tuning.}
\label{Fig3}
\end{figure}
\twocolumngrid\
\vspace*{-0.8cm}
\subsection{E-I reservoirs with inhibitory adaptation have significantly elevated performance}
Our (\( \beta ,\theta\))-parameter analysis shows reservoirs perform best at intermediate global firing rates (Fig.\ \hyperref[Fig1]{1C, D}; \textit{left}). Intermediate firing rates enable neurons to maximize their dynamic range by processing a broad range of inputs \cite{shew_neuronal_2009, ribeiro_deterministic_2008} avoiding periods of silence or saturation. 

We tested whether per-neuron firing rate tuning further enhances RC performance. We implemented a self-adaptation mechanism in which inhibitory connection strengths to a neuron are corrected proportional to the deviation of the neuron’s firing rate from its set target (Fig.\ \hyperref[Fig2]{2A}, Methods). Fig. 2B shows that with the intermediate target firing rate set to exactly 0.5, i. e. homogenously for all neurons, reservoirs initialized with a wide variety of global E/I conditions quickly converge to global balance ($\beta=0$) and rapidly improve in performance. Such performance improvement is even seen for randomly initialized reservoirs that are globally balanced (Fig.\ \hyperref[Fig2]{2B}, \textit{green}), because they too contain local imbalances, which are reduced by the adaptation rule (Fig.\ \hyperref[Fig2]{2B} inset, $n=5$ neurons).

We relaxed the requirement for exact intermediate firing rates using the Beta–distribution which is bounded between 0 (no firing) and 1 (maximal firing) and a symmetrical center at 0.5 to study heterogeneous firing rates (see Methods). This approach allowed convergence to global balance and increased memory capacity across different reservoir initializations, similar to tuning towards homogenous rates. Importantly, variability in the local balance of individual neurons is maintained (Fig.\ \hyperref[Fig2]{2C} inset, n=5 neurons), mirroring cortical firing rate variability. This heterogeneity optimizes RC performance across tasks, as detailed next.
\subsection{Adaptive reservoirs optimize tasks by exploiting different aspects of the sigmoid activation function}
Our self-adapting inhibitory learning boosts RC performance across all tasks (Fig.\ \ref{Fig3}), though the optimal target rate varies by task. With homogeneous target firing rates, memory capacity peaks at 0.5 (\textit{dashed line}), aligning with the linear region of the neuronal sigmoid input/output function (\textit{cf}. Fig. 1A). For other tasks, performance is optimized at slightly higher or lower values, leveraging nonlinear aspects of the sigmoid. Silent ($\sim$0), saturated ($\sim$1) and synchronized (\(\overline{C_{ij}}>0.9\), neither silent nor saturated), states impair performance.

In Fig.\ \hyperref[Fig3]{3B}, we compare the best \textit{non-adaptive} reservoir (gray; optimized via \( \beta\) and \( \theta\)) with two \textit{adaptive} reservoirs (\textit{purple}) with homogeneous (\textit{solid}) and heterogenous (\textit{striped}) target rates, respectively. Adaptive reservoirs improve performance by nearly 100$\%$ over non-adaptive, globally tuned reservoirs. Although heterogenous rates slightly reduce performance, they enhance robustness by eliminating task-specific tuning of the target firing rate.

\begin{figure}[H]
\centering
\includegraphics[width=6.4cm]{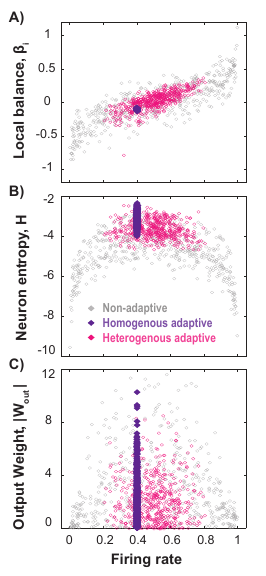}
\caption{\textbf{Adaptive mechanisms enhance RC performance by removing locally imbalanced, low-entropy neurons.
(A)} Scatter plots show that in a globally balanced, non-adaptive RC (\textit{gray}) many neurons exhibit extreme firing rates and local imbalances. An adaptive RC with a homogeneous target rate of 0.4 (\textit{purple}), on the other hand, shifts neurons to the desired rate reducing local imbalances. Heterogeneous targets (\textit{pink}) avoid extreme imbalances while maintaining diverse firing rates. \textbf{(B)} Imbalanced neurons with extreme firing rates (saturated or silent) have low entropy. Adaptation mechanism eliminates these low-entropy neurons for both homogeneous and heterogeneous target rates. \textbf{(C)} Neurons with extreme firing rates and low entropy contribute minimally to predictive performance, as reflected in their low output weights. By avoiding these extreme rates, our inhibitory adaptation mechanism effectively utilizes these neurons, improving predictive performance (see also Fig. 3).}
\label{Fig4}
\end{figure}
\subsection{Adaptive E-I reservoirs enhance performance by eliminating low entropy neurons}
The performance enhancement in RCs achieved through adaptation operates at the level of individual neuron firing rates. Randomly initialized E-I reservoirs, even when globally balanced, contain some neurons that remain locally imbalanced (Fig.\ \hyperref[Fig4]{4A}, \textit{gray}) with nearly saturated ($\sim$1) or silent ($\sim$0) firing rates, making them less responsive to inputs. These neurons show low entropy (Fig.\ \hyperref[Fig4]{4B}, \textit{gray}) and minimal contribution to predictive performance, as indicated by their low output weights (Fig.\ \hyperref[Fig4]{4C}, \textit{gray}).

By applying the local inhibitory adaptation rule with uniform target rates, we can precisely tune the firing rates and, consequently, the local balances (Fig.\ \hyperref[Fig4]{4A}, \textit{purple}). This approach eliminates low-entropy neurons (Fig.\ \hyperref[Fig4]{4B}, \textit{purple}), leading to the elevated performance observed in Fig.\ \hyperref[Fig3]{3A}. Even with heterogeneous target rates (Fig.\ \hyperref[Fig4]{4A}, \textit{pink}), a diverse range of internal states is maintained that avoids extreme firing rates, resulting in significant performance improvements across various tasks (cf. Fig.\ \hyperref[Fig3]{3B}) without the need for fine-tuning rates. We refer to Supp. Fig. 4 for the corresponding analysis with respect to a non-linear task, such as the Lorenz time-series prediction tasks.

\subsection{The adaptive mechanism maximizes performance improvement at each task’s optimal input link scaling, as dictated by the memory-nonlinearity tradeoff}
First documented by Dambre et al. (2012) \cite{dambre_information_2012} and further explored in later studies \cite{inubushi_reservoir_2017, verstraeten_memory_2010, xia_reservoir_2023}, memory-intensive tasks favor more linear reservoir dynamics, whereas tasks like chaotic time-series prediction also benefit from intrinsic nonlinearities. Specifically, in near-linear regimes, the reservoir maps inputs directly into distinct reservoir states, maximizing memory capacity \cite{verstraeten_memory_2010}. This is typically achieved using a small range of input weights, i.e., low input link scaling. In contrast, nonlinear tasks exploit the nonlinear regions of the sigmoid activation function by utilizing a much larger range of input weights, i.e., higher input link scaling.

\begin{figure*}
\centering
\includegraphics[width=17.8cm]{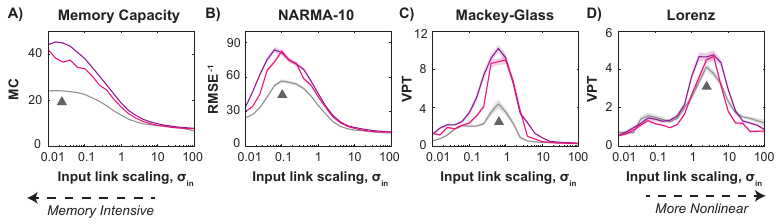}
\caption{\textbf{The adaptive mechanism improves RC performance across input link scaling, with the most significant gains occurring near each task's optimal scaling, determined by the memory-nonlinearity tradeoff.}
Each plot represents a specific task, arranged from memory-dominant (left) to nonlinearity-dominant (right). Colored lines show the mean across 100 runs, with the shaded area indicating the standard error of the mean. Performance of the best non-adaptive, globally tuned system (\textit{gray}) is compared to the adaptive system with homogeneous (\textit{purple}) and heterogeneous (\textit{pink}) target rates as a function of input link scaling, \( \sigma_{in}\). As task nonlinearity increases, the optimal input link scaling for the non-adaptive system ($\blacktriangle$) shifts to higher values, while adaptation consistently improves performance, particularly near each task’s optimal input link scaling.}
\label{Fig5}
\end{figure*}

Our inhibitory adaptation mechanism modifies only internal reservoir weights, allowing us to assess whether a similar trade-off between memory and nonlinearity emerges when adjusting input link scaling for our 4 tasks. In Fig.\ \ref{Fig5}, tasks are arranged from the most memory-intensive (\textit{left}) to the most nonlinear demands (\textit{right}). The memory capacity task performs best at low input link scaling (Fig.\ \hyperref[Fig5]{5A}, $\blacktriangle$), whereas the highly nonlinear Lorenz time-series prediction requires an input link scaling nearly 100 times higher (Fig.\ \hyperref[Fig5]{5D}, $\blacktriangle$). Adaptive reservoirs with homogenous (\textit{purple}) or heterogenous (\textit{pink}) firing rate targets consistently outperform globally tuned reservoirs (\textit{gray}), with the largest performance gains occurring near each task’s optimal input link scaling. The tuned input link scaling, \( \sigma_{in}\), for each task, along with other optimized parameters (\( \beta ,\theta\) for the non-adaptive and \( \rho_{T}\) for the homogenous adaptive method), are provided in the Supp. Table 1.

\vspace{1cm}
\subsection{Designing RCs for specific local balance achieves similar improvements as adaptive mechanisms}
Our adaptation rule in RCs was designed with two main objectives: first, to steer the reservoir toward a specific state of local balance, ensuring the desired distribution of firing rates; and second, to implement a local feedback mechanism that adjusts synapses based on the reservoir’s current state to optimize future behavior. The second objective is inspired by biological systems, which rely on local information for synaptic adjustments.

On the other hand, if the full system and its synaptic weights are known, a reservoir can be explicitly designed with varying degrees of local E/I balance by simultaneously adjusting all inhibitory synapses to achieve a specified target firing rate distribution (see Methods for details). This design, instead of repeatedly applying a local adaptation mechanism, can be directly implemented with either homogeneous (\textit{purple}) or heterogeneous (\textit{pink}) firing rate targets. As shown in Fig.\ \ref{6}, the performance of these designed reservoirs (\textit{striped}) closely aligns with those obtained using the inhibitory adaptation mechanism (\textit{solid}), significantly outperforming the best-performing globally tuned RC, with performance improvements ranging from 15 to 130$\%$. Such designed reservoirs thus offer a straightforward, one-step solution to obtain desired levels of local balance, providing a rapid, scaling advantage for improving globally tuned, randomly initialized reservoirs.
\onecolumngrid\
\begin{figure}[h]
\centering
\includegraphics[width=18cm]{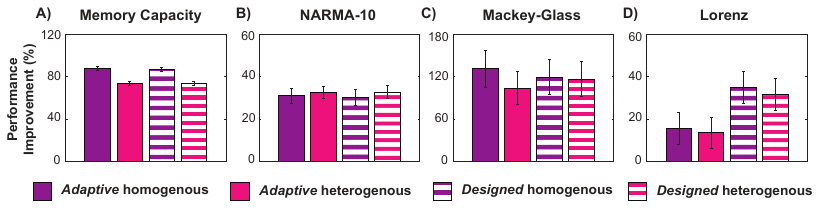}
\caption{\textbf{One-step designed random reservoirs with homogeneous or heterogeneous target rates achieve performance gains similar to using local balance through adaptative convergence.}
Designed reservoirs (\textit{striped}) show performance improvements comparable to those achieved by the local inhibitory adaptation mechanism (\textit{solid}), whether using homogeneous (\textit{purple}) or heterogeneous (\textit{pink}) firing rate targets. Both approaches yield up to a 130$\%$ improvement in performance compared to a non-adaptive, globally tuned reservoir. \textbf{(A) }Memory Capacity: MC increase. \textbf{(B)} NARMA-10: RMSE decrease. \textbf{(C, D)} Mackey-Glass and Lorenz time series: VPT increase.}
\label{Fig6}
\end{figure}
\newpage
\twocolumngrid\
\vspace{-1.1cm}
\section{Discussion}
In this study, we investigate how tuning the excitatory-inhibitory balance in neural reservoirs can lead to significant performance improvements. We go beyond the generic global balance seen in typical RC implementations by probing the role of local balance in a brain-inspired excitatory inhibitory (E/I) reservoir. Specifically, we introduce a novel inhibitory adaptation mechanism that rapidly steers the network to appropriate levels of both local and global E/I balance to reach a high performing dynamical regime. This self-tuning, coupled with variability in target firing rates observed in biological neural networks, eliminates the need for extensive external fine-tuning and boosts performance across diverse tasks such as memory capacity and nonlinear time series prediction. Furthermore, we demonstrate that networks can be explicitly designed to reach desired levels of local balance through a single-step adjustment of randomly initialized inhibitory links, providing a more efficient and scalable alternative to iterative tuning.

\subsection*{The role of excitation-inhibition balance in reservoir computing}
Excitatory-inhibitory (E-I) balance is a fundamental principle in neural systems, yet its role in reservoir computing has received relatively little attention \cite{kanamaru_maximal_2023, de_graaf_increased_2022}. Traditional reservoir computing approaches often use a hyperbolic tangent (tanh) activation function, which obscures the effects of E-I dynamics. Because neurons can output both positive and negative values, and synaptic weights can also be either sign, the distinction between excitation and inhibition becomes ambiguous. As a result, a form of global balance typically emerges simply from the random initialization of synaptic weights, rather than through any structural feature of the reservoir.

In our study, we explicitly separated excitatory and inhibitory dynamics by employing a sigmoid neuron-like activation function and defining distinct populations of excitatory and inhibitory neurons. This separation allowed us to systematically explore how different E-I regimes shape reservoir dynamics and computational performance. We found that reservoirs operating in balanced to slightly inhibition-dominated regimes exhibit enhanced performance across tasks. Inhibitory neurons play a crucial role in preventing runaway excitation, ensuring that the network remains within a dynamical regime that supports efficient information processing.

This asymmetrical preference toward inhibition is consistent with prior findings by deGraaf et al. \cite{de_graaf_increased_2022}, who showed that inhibition-dominated networks maintain balanced neuronal input and exhibit diverse firing rate variability, whereas excitation-dominated networks suffer from saturated firing rates. The latter reduces the effective dimensionality of network activity, leading to overly simplified output patterns. In line with these findings, our results suggest that while reservoirs can tolerate excess inhibition, increased excitation degrades performance—highlighting a fundamental asymmetry in how neural networks process information.

The mechanisms governing E-I balance in biological brains are complex, ranging from anatomical constraints—such as fixed excitatory-to-inhibitory ratios—to dynamic synaptic adjustments that actively maintain equilibrium. One such anatomical constraint is Dale’s Law, which states that a given neuron releases either excitatory or inhibitory neurotransmitters, but not both \cite{kandel_cellular_1968, strata_dales_1999}. While this principle is crucial in biological systems, our results suggest it is not essential for effective reservoir computation. Even when we shuffled synaptic weights, breaking the strict segregation of E and I neurons, we observed similar computational performance (see Supp. Fig. 3). This principle likely evolved as a consequence of biological constraints during neural development, where excitatory and inhibitory neurons develop distinct morphologies and neurotransmitter systems. However, further study is warranted to better understand the role of distinct excitatory and inhibitory populations in shaping network dynamics and computational properties. Notably, we find that strong RC performance in balanced and slightly over-inhibited regimes remains consistent whether E–I balance is tuned by adjusting the strength of inhibitory connections or by varying the proportion of inhibitory neurons (see Supp. Fig. 1), suggesting flexibility in how balance can be implemented in artificial systems.

To reach an appropriate level of local balance in our reservoir computing models, we introduced an inhibitory adaptation rule that dynamically tunes inhibitory synapses at the neuronal level. While this rule draws inspiration from biological principles of firing rate homeostasis, it is not intended as a direct model of neural E-I regulation. Instead, it serves as a computationally efficient mechanism for improving reservoir function by leveraging biologically motivated constraints.

\subsection*{Adaptive mechanisms make reservoirs more scalable by alleviating hyperparameter optimization}
Because typical RCs rely on generic randomly constructed neural reservoirs, incorporating bio-inspired principles offers a promising path for improving both performance and robustness. Prior research has explored various modifications to network topology of the reservoir, such as small-world connectivity and structured recurrent architectures, to enhance computational capacity \cite{pathak_hybrid_2018,rodriguez_optimal_2019,srinivasan_parallel_2022,gallicchio_deep_2017}. However, most RC optimization strategies remain focused on task-specific performance and achieve it through extensive hyperparameter tuning. This process, often the most computationally demanding step in RC implementations, requires multiple training cycles to evaluate different configurations. The computational cost of each training step scales cubically with network size—O(N³) in typical implementations, where N is the number of neurons, making exhaustive optimization infeasible for large-scale systems. Even simple grid searches become prohibitively expensive when optimizing just a few hyperparameters, and while more sophisticated techniques like Bayesian optimization \cite{yperman_bayesian_2016,antonik_bayesian_2023} or genetic algorithms \cite{ferreira_approach_2013} can improve efficiency, they still require numerous costly training iterations.

To overcome this limitation in RCs, we propose shifting from performance-based \textit{optimization} to dynamics-based \textit{adaptation}. Rather than searching the high-dimensional parameter space for optimal configurations, our local adaptation rule enables reservoirs to self-organize into dynamical regimes that naturally support high performance. This approach significantly reduces the need for global hyperparameter tuning, making larger reservoir implementations computationally feasible. Additionally, we introduce a complementary non-local design rule that provides an efficient, one-step alternative, particularly useful for large-scale implementations where local adaptations, though far less costly than typical hyperparameter optimization, may still introduce undesired computational overhead.

Many prior adaptation mechanisms \cite{babinec_improving_2007,schrauwen_improving_2008,morales_unveiling_2021,yusoff_modeling_2016} have demonstrated performance improvements in RCs, and our local E-I balance-based rule is one such approach. However, the important takeaway is not the specific implementation of this rule but the broader paradigm shift—from statically \textit{optimized} reservoirs to dynamically \textit{adaptive} systems. This perspective highlights a promising direction for addressing scalability challenges and improving the efficiency and robustness of RC implementations.

\subsection*{Firing rate heterogeneity allows reservoirs to adapt to task-specific demands}
While adaptation reduces the need for manual tuning of multiple reservoir parameters, it introduces a new consideration: determining the optimal target firing rate. This target is inherently task-dependent due to a fundamental trade-off between linear and nonlinear processing needs. Neurons operating largely linear, i.e., in the middle, of their activation range ($\sim$0.5 firing rate), maintain maximal sensitivity to input history, enhancing memory. Conversely, neurons operating inclusively towards non-linear aspects of the input-output function, exhibit more nonlinear responses, improving the network's ability to perform complex transformations.

Our results confirm this trade-off: memory-intensive tasks, such as memory capacity, perform best when neurons maintain intermediate firing rates and corresponding local balance between excitation and inhibition. In contrast, highly nonlinear tasks, like chaotic time-series prediction, benefit from deviations that enhance nonlinearity. This task dependence creates a challenge when designing adaptable RCs without reintroducing task-specific tuning of reservoir weights.

To address this, we incorporate another biologically inspired principle: neuronal firing rate heterogeneity. In biological systems, distinct populations of neurons maintain different baseline firing rates, allowing them to fulfill diverse computational roles. Prior research has shown that combining linear and nonlinear nodes enhances multitask performance \cite{inubushi_reservoir_2017}. Importantly, we find that balanced reservoirs with heterogeneous firing rate targets also naturally balance memory and nonlinearity, yielding broad performance gains while eliminating task-specific tuning of firing rates. In these reservoirs, some neurons optimize memory retention, while others enhance nonlinear processing, reflecting the division of labor observed in biological neural circuits. This heterogeneity provides a flexible and robust computational architecture, enabling reservoirs to generalize more effectively across diverse tasks.

By combining E/I balance with firing rate heterogeneity, our study highlights how biologically inspired mechanisms can enhance artificial neural computation. Our results demonstrate that appropriately balancing excitation and inhibition at the local level improves computational efficiency, while incorporating heterogeneity makes reservoirs more flexible across tasks. Moreover, by reducing reliance on hyperparameter optimization, our approach significantly enhances scalability, making it well-suited for larger and more complex learning systems. As AI models continue to grow in size and computational demand, biologically inspired adaptation mechanisms may offer a crucial step toward more efficient, self-organizing, and scalable neural architectures.
\section{Acknowledgements}
This research was supported by the Division of the Intramural Research Program (DIRP) of the National Institute of Mental Health (NIMH), USA, ZIAMH002797, ZIAMH002971, and the BRAIN initiative Grant U19 NS107464-01 to D. P. M.G.’s contributions were supported by ONR Grant N000142212656. This research utilized the supercomputing resources of the National Institutes of Health (NIH, USA; Biowulf, \url{http://hpc.nih.gov}) and the University of Maryland (UMD College Park, USA, \url{https://hpcc.umd.edu/hpcc/zaratan.html}).
\section{Author Contributions}
K.S., D.P., and M.G. conceived the study; K.S. did the simulations and analysis; M.G. and D.P. acquired funding and supervised the study; K.S., D.P., and M.G. wrote the manuscript. 
\section{Ethics declarations}
The authors declare no competing interests.
\section{Appendix}
\subsection{Constructing the adjacency matrix,}
In the Methods section, we outline the construction of a recurrent circuit with a tunable excitatory-inhibitory (E-I) balance. The network's adjacency matrix, \( A\), is divided into two submatrices: \( A_{E}\) for excitatory links and \( A_{I}\) for inhibitory links. Neurons are classified as excitatory or inhibitory, with each neuron’s outgoing links exclusively matching its type, adhering to Dale's Law. The network’s average degree is higher than in typical RC setups, for two key reasons: (i) to reflect the brain's high degree of neuronal connectivity, and (ii) to avoid neurons with only one or two inhibitory connections, which can lead to frustrated arrangements during inhibitory adaptation. The non-zero entries in the excitatory submatrix are drawn from a normal distribution with a mean \( \mu_{E}\) given by \( 1/(d\ast f_{E}),\) where \( d\) is the average degree and \( f_{E}\) is the fraction of excitatory neurons. This is done so that the average incoming excitatory current to a neuron is 1. The standard deviation of this distribution is \( \sigma_{E} = 0.2\mu_{E}\), or 20$\%$. The non-zero entries in the inhibitory submatrix are drawn from a different normal distribution with a mean \( \mu_{I}\), which is varied to control the relative E-I balance in the system. The mean \( \mu_{I}\) is defined by the following expression:

\begin{equation}
\mu_{I} =\frac{f_{E}\cdot\mu_{E}-\beta /k}{1-f_{E}}
\end{equation}

Here, \( \beta\) is the global balance parameter, with \( \beta = 0\) indicating global balance. The standard deviation of the inhibitory link distribution, \( \sigma_{I}\), is the same as that of the excitatory link distribution, \( \sigma_{E}\).

\subsection{Heterogeneous distribution of firing rate targets}
In the Methods and Results sections, we refer to the heterogeneous distribution of firing rate targets. This section provides detailed information on the subject. Figure \ \hyperref[Fig3]{3A} illustrates that neurons with extreme firing rates (\( \sim 0\) or \( \sim 1\)) exhibit low entropy and contribute less to decision-making processes. These neurons are effectively "wasted." In designing a heterogeneous distribution, we aimed to satisfy two criteria: i) limit the range of neuronal firing rates to avoid extreme values, and ii) maintain a reasonable fraction of neurons near 0.5 while preserving a spread of firing rates to exploit their nonlinearity.

The \( Beta\)-distribution proved suitable for this purpose. Since our firing rate values are confined between 0 and 1, the \( Beta\)-distribution is a natural fit. For the parameters of the distribution (\( Beta(a,b)\)), we set \( a = b\) to ensure a symmetric distribution with a mean of 0.5. We chose the parameter value of 9 to achieve a range that avoids excessively high or low values. Consequently, for all analyses involving a heterogeneous distribution of firing rate targets, we utilize the \( Beta(9,9)-\)distribution. However, we expect other distributions that meet the criteria mentioned to yield comparable results. 

\subsection{Non-parametric estimator for differential entropy}
This estimator, first introduced by Kozachenko and Leonenko \cite{kozachenko_sample_1987}, is given by the following equation:

\begin{equation}
 H\left(r_{i}\right) = \psi\left(T\right)-\psi\left(1\right)+\text{log}\left(2\right)+\frac{1}{T}\sum_{t = 1}^{T}\text{log }\epsilon_{i}\left(t\right)
\end{equation}
Here, \( H\left(r_{i}\right)\) is the entropy estimate of the state of the \( i^{\text{th}}\) neuron. \( \psi\) is the digamma function, \( T\) is the total number of steps used to estimate the entropy, which in our case was \( 10^{4}\), and \( \epsilon_{i}\left(t\right)\) is the nearest neighbor distance of \( r_{i}\left(t\right)\). Unlike Shannon entropy, commonly used for discrete values, differential entropy can be negative.
\newpage
\bibliography{BrainRC.bib}

\end{document}


\title{Supplementary Material\\ Boosting Reservoir Computing with Brain-inspired Adaptive Dynamics}
\author{Keshav Srinivasan}
\affiliation{Biophysics Program, University of Maryland, College Park, MD 20740, USA}
\affiliation{Section on Critical Brain Dynamics, National Institute of Mental Health, Bethesda, MD 20892, USA}
\author{Dietmar Plenz}
\affiliation{Section on Critical Brain Dynamics, National Institute of Mental Health, Bethesda, MD 20892, USA}
\author{Michelle Girvan}
\affiliation{Biophysics Program, University of Maryland, College Park, MD 20740, USA}
\affiliation{Department of Physics, University of Maryland, College Park, MD 20740, USA}
\affiliation{Santa Fe Institute, Santa Fe, NM 87501, USA}

\maketitle

\section*{Supplementary Note: Calculation for
'Designed' reservoirs}
Consider the 2 reservoir equations (refer to Equations 1 and 2 from the
main text):
\begin{equation}
V_{i}(t + 1) = \lambda_{i}V_{i}(t) + \sum_{j = 1}^{N}\mspace{2mu}\left( A_{ij}^{E} - A_{ij}^{I} \right) \cdot r_{j}(t) + W_{i}^{in}u(t) 
\end{equation}
\begin{equation}
r_{i}\,(t) = \text{Sig}\,\left( V_{i}\,(t) - \theta_{i}\,(t) \right)
\end{equation}
As mentioned in the main text, here \(V_{i}(t)\) is the membrane
potential of the \(i^{\text{th}}\) neuron, \(\lambda\) is the leakage term (which in this case is set to 0), \(A\) is the connectivity matrix of the reservoir network, \(W^{in}\) is the input matrix and \(u(t)\) is the input data. The second equation defines the reservoir variable, \(r(t)\), which represents the firing rate of the neurons in the network and can take values between 0 and 1. Sig is a sigmoid function and is given the following equation:
\begin{equation}
\text{Sig}(y) = \frac{1}{1 + e^{- 10y}}
\end{equation}

Consider a steady-state solution with a firing rate set-point of
\(\rho\). We can now rewrite the steady-state version of Equation (1)
as:
\begin{equation}
{\widetilde{V}}_{i} = \sum_{j = 1}^{N}\mspace{2mu}{\widetilde{A}}_{ij}\rho_{j} + W_{i}^{in}\langle u\rangle
\end{equation}

Inverting equation 2 for the steady state solution, we also obtain:
\begin{equation}
{\widetilde{V}}_{i} = \text{Logit}(\rho_{i}) + \theta_{i}
\end{equation}

Here Logit is the inverse Sigmoid function. Combining equations 4 and 5 we then get:
\begin{equation}
\sum_{j = 1}^{N}\mspace{2mu}{\widetilde{A}}_{ij}\rho_{j} = \text{Logit}(\rho_{i}) + \theta_{i} - W_{i}^{in}\langle u\rangle
\end{equation}

Let us consider a design mechanism that only alters the inhibitory links by multiplying them by a neuron-wise multiplicative factor, \(\Omega_{i}\).

Before the adjustment, we have the following equation (Split into
excitatory and inhibitory sub-parts):
\begin{equation}
\sum_{j = 1}^{N}\mspace{2mu} A_{ij}\rho_{j} = \sum_{E}^{}\mspace{2mu} A_{ij}\rho_{j} + \sum_{I}^{}\mspace{2mu} A_{ij}\rho_{j}
\end{equation}

And after the adjustment, we get the following equation:
\begin{equation}
\sum_{j = 1}^{N}\mspace{2mu}{\widetilde{A}}_{ij}\rho_{j} = \sum_{E}^{}\mspace{2mu} A_{ij}\rho_{j} + \Omega_{i}\sum_{I}^{}\mspace{2mu} A_{ij}\rho_{j}
\end{equation}

Combining Eqns 6 and 8 we then finally get an expression for the
multiplicative factor, \(\Omega_{i}\):
\begin{equation}
\boxed{\Omega_{i} = \frac{\text{Logit}(\rho_{i}) + \theta_{i} - W_{i}^{in}\langle u\rangle - \sum_{E}^{}\mspace{2mu} A_{ij}\rho_{j}}{\sum_{I}^{}\mspace{2mu} A_{ij}\rho_{j}}}
\end{equation}

\section*{Supplementary Figures}

\begin{figure}[htp!]
\centering
\includegraphics[width=15cm]{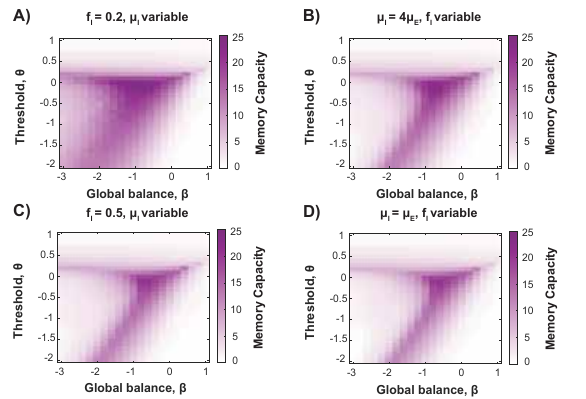}
\hfill
\caption{\textbf{Reservoir performance remains robust across different forms of E-I balance modulation.} Memory capacity is shown as a function of threshold ($\theta$) and global balance ($\beta$). We vary global balance in two ways: by changing the mean strength of inhibitory links (\textbf{A, C}) and by changing the fraction of inhibitory neurons (\textbf{B, D}). In (\textbf{A}) and (\textbf{C}), the proportion of inhibitory neurons is fixed (20\% and 50\%, respectively), In (\textbf{B}) and (\textbf{D}), the E/I synaptic strength ratio is fixed (1:4 and 1:1, respectively). Across all conditions, high performance is consistently observed in balanced to slightly inhibited regimes, demonstrating the robustness of reservoir dynamics to the method of E-I tuning.}
\end{figure}

\begin{figure}[htp!]
\centering
\includegraphics[width=18.2cm]{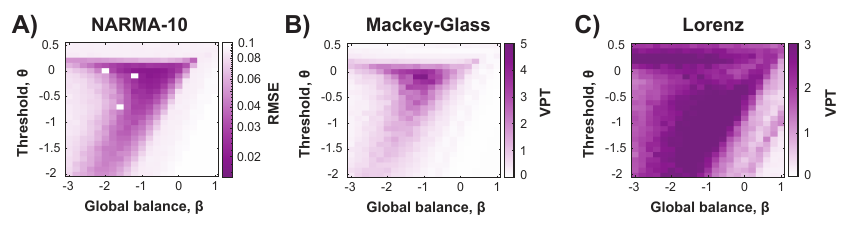}
\hfill
\caption{\textbf{Performance as a function of the balance parameter $\beta$ and threshold $\theta$ for three time-series prediction tasks: NARMA-10, Mackey-Glass, and Lorenz.}
This figure expands on Fig. 1C in the main text, offering a detailed analysis of performance trends across these tasks. Performance remains high and stable in balanced or slightly over-inhibited states but declines sharply in the over-excited regime, where it becomes more fragile across all three tasks.
}
\end{figure}

\begin{figure}[H]
\centering
\includegraphics[width=15cm]{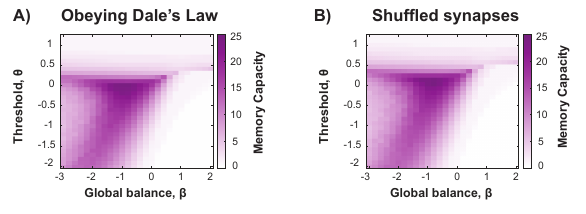}
\hfill
\caption{\textbf{Reservoir dynamics and performance remains consistent despite synapse shuffling, indicating no strong dependence on Dale’s Law.}
(\textbf{A}) A globally tuned excitatory-inhibitory (E-I) reservoir where neurons and synapses strictly follow Dale’s Law (as shown in Fig. 1D). (\textbf{B}) A modified reservoir where synapses are randomly shuffled, disrupting Dale’s Law while preserving overall connectivity statistics. Despite this alteration, performance remains unchanged across different values of the global balance parameter ($\beta$) and threshold ($\theta$), demonstrating that adherence to Dale’s Law is not a necessary condition for RC performance
}
\end{figure}
\newpage
\begin{figure}[H]
\centering
\includegraphics[width=18cm]{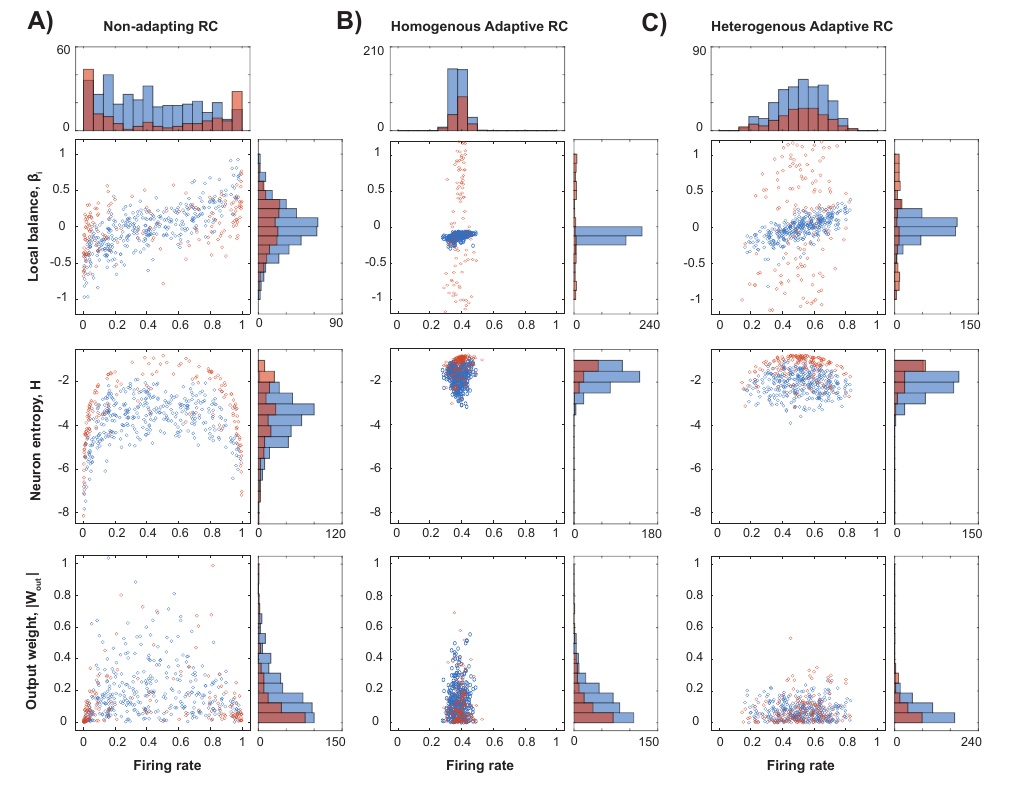}
\hfill
\caption{\textbf{Scatter plots depicting the relationship between local balance, neuronal entropy, and output weights as functions of average firing rate for input and non-input neurons during a Lorenz time-series prediction task.} (\textbf{A}) In the scatter plots of a globally balanced, non-adapting reservoir computer (RC) ($\beta=0$), we observe (\textit{top}) that input neurons (\textit{orange}) are driven more strongly, resulting in higher firing rates compared to non-input neurons (\textit{blue}). (\textit{middle}) The neurons with extreme firing rates show reduced entropy, which (\textit{bottom}) leads to their diminished contribution to prediction accuracy. (\textbf{B}) When an adaptive RC with a homogeneous firing rate target of 0.4 is employed, (\textit{top}) the adaptation mechanism tunes the network to achieve the desired firing rate. Here, input neurons exhibit substantial local imbalance to compensate for the strong input signals. (\textit{middle}) This adaptation process successfully eliminates low-entropy neurons, (bottom) ensuring that all neurons contribute adequately to predictions. (\textbf{C}) As in the previous case, with heterogeneous targets, (\textit{top}) input neurons require significant local imbalance to adjust for the strong input. By mitigating extreme firing rates, (\textit{middle}) we also eliminate low-entropy neurons, (\textit{bottom)} resulting in a trend in output weights that parallels the findings with homogeneous targets.}
\end{figure}
\newpage
\section*{Supplementary Table}
\begin{table}[ht]
\centering
{\renewcommand{\arraystretch}{2}
\begin{tabularx}{0.8\textwidth}{|>{\centering\arraybackslash}p{3cm}|*{6}{Y|}}
\hline
\multirow{3}{*}[0pt]{\centering \textbf{Task}} & \multicolumn{3}{c|}{\multirow{2}{*}{\textbf{Non-adaptive}}} & \multicolumn{3}{c|}{\textbf{Adaptive}} \\ \cline{5-7}
 & \multicolumn{3}{c|}{} & \multicolumn{2}{|c|}{\textbf{Homogenous}} & \textbf{Hetero.}\\ \cline{2-7}
              & $\sigma_{in}$ & $\beta$ & $\theta$ & $\sigma_{in}$ & $\rho_T$ & $\sigma_{in}$ \\
\hline
Memory Capacity & 0.016 & -1.0 & 0.0 & 0.016 & 0.5 & 0.010 \\
\hline
NARMA-10        & 0.100 & -1.0 & 0.0 & 0.063 & 0.3 & 0.100 \\
\hline
Mackey Glass    & 0.631 & -1.2 & 0.0 & 0.631 & 0.3 & 1.000 \\
\hline
Lorenz          & 2.512 & -1.0 & -1.0 & 3.981 & 0.6 & 3.981 \\
\hline
\end{tabularx}
}
\caption{Task-dependent parameters used in the RC setup}
\end{table}